\documentclass[10pt,twocolumn,letterpaper]{article}

\usepackage{iccv}
\usepackage{times}
\usepackage{epsfig}
\usepackage{graphicx}
\usepackage{amsmath}
\usepackage{amssymb}
\usepackage{float}
\usepackage{graphicx}
\usepackage{subcaption}
\usepackage{rotating}
\usepackage{lipsum} 

\rmfamily

\usepackage[breaklinks=true,bookmarks=false]{hyperref}

\iccvfinalcopy 

\def\iccvPaperID{12078} 

\ificcvfinal\fi

\begin{document}

\title{CHATTY: Coupled Holistic Adversarial Transport Terms with Yield for Unsupervised Domain Adaptation}

\author{Chirag P, Mukta Wagle\\
Department of Electrical Engineering\\
Indian Institute of Technology Bombay\\
Mumbai, India\\
{\tt\small 18B090003@iitb.ac.in}\\
{\tt\small 18D070054@iitb.ac.in}
\and
Ravi Kant Gupta\\
Department of Electrical Engineering\\
Indian Institute of Technology Bombay\\
Mumbai, India\\
{\tt\small 184070025@iitb.ac.in}
\and
Pranav Jeevan P\\
Department of Electrical Engineering\\
Indian Institute of Technology Bombay\\
Mumbai, India\\
{\tt\small 194070025@iitb.ac.in}
\and
Amit Sethi\\
Department of Electrical Engineering\\
Indian Institute of Technology Bombay\\
Mumbai, India\\
{\tt\small asethi@iitb.ac.in}\\
}


\maketitle
\ificcvfinal\thispagestyle{empty}\fi

\begin{abstract}
   
We propose a new technique called CHATTY: Coupled Holistic Adversarial Transport Terms with Yield for Unsupervised Domain Adaptation. Adversarial training is commonly used for learning domain-invariant representations by reversing the gradients from a domain discriminator head to train the feature extractor layers of a neural network. We propose significant modifications to the adversarial head, its training objective, and the classifier head. With the aim of reducing class confusion, we introduce a sub-network which displaces the classifier outputs of the source and target domain samples in a learnable manner. We control this movement using a novel transport loss that spreads class clusters away from each other and makes it easier for the classifier to find the decision boundaries for both the source and target domains. The results of adding this new loss to a careful selection of previously proposed losses leads to improvement in UDA results compared to the previous state-of-the-art methods on benchmark datasets. We show the importance of the proposed loss term using ablation studies and visualization of the movement of target domain sample in representation space.

\end{abstract}

\section{Introduction}

\begin{figure}[h]
\begin{center}
   \includegraphics[width=0.95\linewidth]{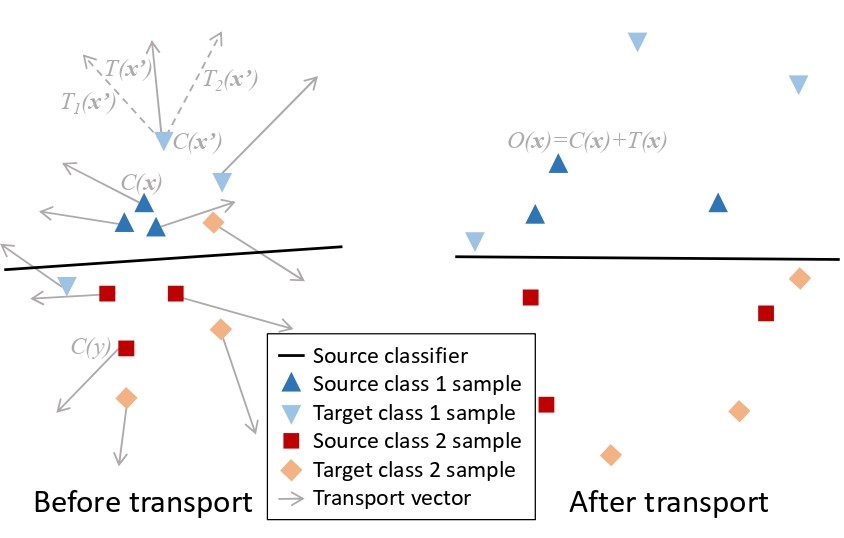}
\end{center}
   \caption{UDA for classification can be achieved by shifting the samples in a way that favors finding a common decision boundary for source and target domains. See text for details}
\label{fig:concept1}
\end{figure}

Domain shift is a practical problem faced by machine learning models in the real-world, where the distribution of the test data (roughly, \emph{target} domain) is different from that of the main component of the training data (roughly, \emph{source} domain), which causes a drop in model generalization accuracy on the former. Domain adaptation (DA) is a sub-field of machine learning that focuses on developing algorithms and techniques to train models that effectively transfer knowledge learned from a source domain to a target domain~\cite{da}. The DA problem may be cast as one of learning to represent the source and the target domain samples in common feature space where their distributions overlap and the subsequent part of the model is able to perform the task with good generalization on both domains~\cite{ondeep}. DA techniques can be further classified based on various assumptions about the extent of availability of data from the target domain during training. For instance, semi-supervised DA assumes availability of some labeled samples and lots of unlabeled samples, few-shot DA assumes some labeled samples and no unlabeled samples, unsupervised DA (UDA) assumes only unlabeled samples, and domain generalization assumes that no sample is available from the target domain during training.  For a classification task, if the classes in the source and target domains are the same, it is called closed-set domain adaptation.  We provide a novel solution for UDA for the closed-set. UDA is instrumental in training transferable models which are pre-trained on a different dataset, without having to worry about labeling any number of images from the target domain. It has major applications in various areas, including medical image analysis, where the image acquisition setup (e.g., equipment, technician training and protocols) is often different between model development and deployment scenarios.

There are some area-specific UDA techniques, such as color normalization for pathology images~\cite{vahadane,fastgpu}, but we are interested in a more general solution. A common approach for a general UDA technique is to use a moment matching algorithm, which focuses on aligning feature distributions of the source and the target domains by matching their moments~\cite{mmd,mdd,deepcoral,homm,dcan}. However, more successful approaches to deep domain adaptation have been those that train networks to extract feature that confuse a domain discriminator neural network in an adversarial fashion. Such neural networks extract domain-invariant representations, i.e., those cannot be used to distinguish between examples from the source and target domains~\cite{dann,cdan}. We propose a novel scheme to \emph{transport} samples in the classification (logit) space for adversarial UDA (see Figure~\ref{fig:concept1}).

CHATTY, or Coupled Holistic Adversarial Transport Terms with Yield, for unsupervised domain adaptation, is our proposed approach. Our architecture is adversarial by nature~\cite{advda}. We shift each final representation of the source and target samples by two learnable and sample-dependent displacement vectors, which we call \emph{transport} terms. We train a feature extractor in a way that the two transport terms displace different samples in different directions. We also take care that the two transport terms move similar samples similarly. This is done by a bilinear function that we call the transport loss, which has the class information optionally embedded into the loss matrix.

We evaluate the performance of CHATTY on Office-Home~\cite{office-home}, Office-31~\cite{office31}, and a medical image dataset -- FHIST~\cite{fhist}. Medical image classification is a critical yet challenging application of deep domain adaptation as image acquisition setups often change. Moreover, medical images often do not contain an object of invariance as opposed to natural images. This makes domain adaptation tasks significantly harder. Class confusion is a rampant problem in medical image datasets, owing to the near identical features of images of different classes~\cite{mcc}. We posit that reducing class confusion in medical image domain adaptation tasks is crucial, and CHATTY when optionally combined with class confusion minimization solves this problem well.

To summarize, our main contributions are the following:
\begin{enumerate}
    \item Reducing the confusion in classification by transporting the samples to a different region of the classification space,
    \item Formulating a loss term to move similar samples in similar directions using two transport terms,
    \item Designing an end-to-end network which uses both the source and the target datasets judiciously by combining the proposed loss with other useful losses, and compares favorably to the state-of-the-art CNN DA architectures,
    \item Producing a detailed analysis of the transport loss, the MCC loss \cite{mcc} and their effects in our experiments on three datasets: OfficeHome~\cite{office-home}, Office-31~\cite{office31} and a medical image dataset, FHIST~\cite{fhist}, and
    \item Visualizing the effects of the transport loss in the classification space.
\end{enumerate}


\section{Background and Related Works}

In UDA, we have the source domain samples along with their labels $D_s = \{(x_i^s,y_i^s)\}_{i=1}^{N_s}$ where the number of classes is $C$. We also have the unlabeled target domain samples $D_t = \{x_i^t\}_{i=1}^{N_t}$. Most contemporary UDA techniques train neural networks end-to-end with multiple loss functions, instead of training multiple networks separately~\cite{mmd,homm,deepcoral,dcan,cdan,srdc}. One group of losses for end-to-end training tries to improves classification accuracy on the labeled samples (source domain), such as by using the cross entropy given as: 
\begin{equation}
    \mathcal{L}_{c} = \frac{1}{N_s} \sum_{i=1}^{N_s}L_{ce}(C(G(x_i^s)),y_i^s),
\end{equation}
where $G$ is a feature extractor sub-network, $C$ is a classifier sub-network, and $L_{ce}$ is the cross entropy loss. The other group of losses tries to match the distributions of representations of the source and target domain samples. Additional loss terms may be used that try to keep unlabeled samples (target domain) well-separated near classification decision boundaries.


Moment matching methods, such as MMD~\cite{mmd}, MDD~\cite{mdd} and DCAN~\cite{dcan}, aim to match expectations (means) and higher moments of the features of samples of the source and the target domains. However, these techniques do not solve the subtle problem of feature confusion, where one feature is confused for the another between the two domains. For example, the second moments of two features that are semantically different between the two domains could be equal, and yet correlation alignment would align incompatible features. Deep CORAL~\cite{deepcoral} and HoMM~\cite{homm} propose matching additional measures of the distributions, such as the covariance matrix, to solve this problem.

Adversarial methods take another approach that aims to confuse an additional network on top of the feature extractor, called the discriminator, whose aim is to discriminate between the two domains. The confusion is formulated in terms of an adversarial loss. The two representative methods of this class are DANN~\cite{dann} and CDAN~\cite{cdan}. CDAN is an improved version of DANN because it captures the relationships between the parameters of the feature extractor and the classifier, rather than just treating them independently. SRDC uses a clustering algorithm to group similar data points together, and finally optimizes the deep representation learning using a joint objective that combines clustering loss and domain classification loss~\cite{srdc}. 

There are also certain hybrid techniques, which use a combination of moment matching and adversarial training concepts. SHOT projects the data onto a set of random directions, computes the 1D Wasserstein distances between the projected distributions, and optimizes the projections to minimize the Wasserstein distance~\cite{shot}. SymNets uses three classifiers, one specifically for the source domain, one specifically for the target domain and one for both, and seeks to minimize classification loss on all of them~\cite{symnets}. CyCADA maximizes the cycle consistency which tracks the feature distribution of the image and enforces that the distributions after a cyclical transform are equal~\cite{cycada}. GVB-G is a theoretically sound method, which directly minimizes the extent of the discrepancies by minimizing the length of certain bridge terms between the two domains~\cite{gvb}.  

Some methods also aim to make the classifier more robust in order to learn invariant feature representations. Maximum Classifier Discrepancy (MCD) aims to learn different representations of the source and target, and learning is made more robust by increasing the discrepancies between the two classifiers in terms of the features they use to classify~\cite{mcd}. Minimum Class Confusion (MCC) aids the classifier to make more confident predictions by minimizing the class confusion probabilities, and in doing so, reducing the number of samples being represented very close to the decision boundaries in the classification space~\cite{mcc}.


\section{Proposed Approach: CHATTY}

As mentioned in the previous section, the two objectives of several successful UDA methods are to improve the classification accuracy on labeled samples (source domain) and to increase the overlap between the feature distributions of the two domains. 

\begin{figure*}
\begin{center}
   \includegraphics[width=1.05\linewidth,trim= 0 50 0 0,clip]{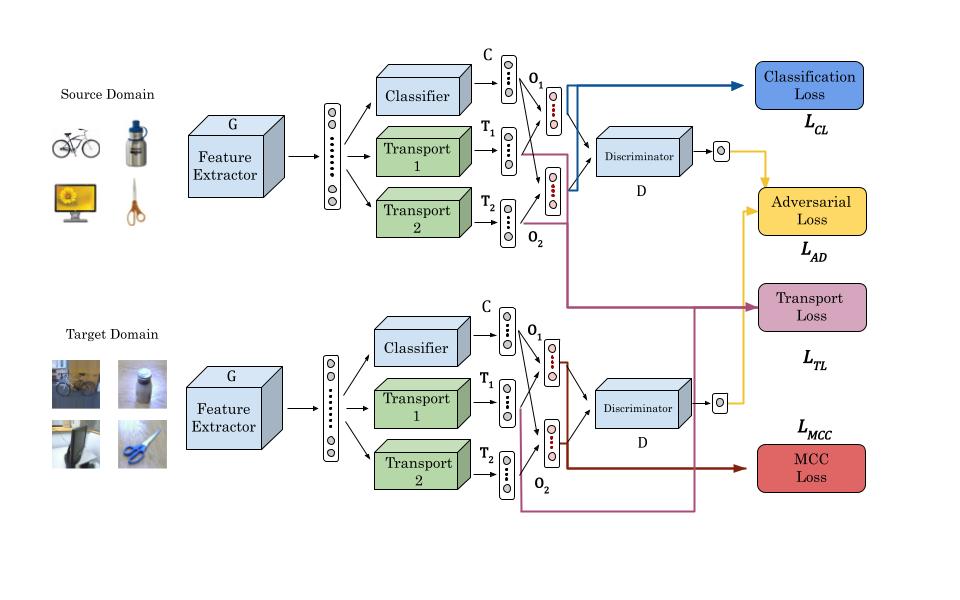}
\end{center}
   \caption{The proposed CHATTY framework, which is shared by the source and the target domain uses Transport Layers $\mathbf{T_1}$ and $\mathbf{T_2}$ compute the transport vectors that move classifier outputs to their final locations.}
\label{fig:framework}
\end{figure*}

There are various ways to increase the overlap between the two distributions, as discussed above, but we propose a different approach since none of the prior UDA techniques seem to have tried a transport approach that spread the sample representations out using learnable displacements. The objectives of such a spread would be to ease the crowding around classification boundaries (minimize class confusion~\cite{mcc}, but in a different way), \emph{and} to align the distributions of the source and target domains~\cite{mmd,homm,deepcoral,dcan,cdan,srdc}, Additionally, there seems to be some scope left to combine useful aspects of these disparate types of approaches to UDA. These observations summarize the motivation behind the proposed CHATTY framework for UDA. Figure~\ref{fig:framework} shows a visual description for our approach. Different components of this framework are described in each subsection below.

\subsection{Overall objective}

The general problem of adversarial domain adaptation (ADDA)~\cite{adda} for classification can be formulated as follows:

\begin{equation}
\begin{aligned}
    \max_D \mathcal{L}_{adv}(G, C, D) + \mathcal{L}_c(G,C) + \mathcal{L}_e(G),
\end{aligned}
\end{equation}

\begin{equation}
\begin{aligned}
    \min_{G,C} \mathcal{L}_{adv}(G, C, D) + \mathcal{L}_c(G,C) + \mathcal{L}_e(G), 
\end{aligned}
\end{equation}
where $G$ is the feature extractor, $C$ is the classifier, and $D$ is the domain discriminator, $\mathcal{L}_{adv}$ is the adversarial loss, and $\mathcal{L}_e$ is the external loss as given in~\cite{adda}. While a classifier head and loss may drive the labeled source samples from different classes far in the classifier (logit) space, without other objectives, such as $\mathcal{L}_{adv}$, the samples of the target domain may fall close to the decision boundaries. We seek to reduce the number of samples falling very close to the decision boundaries for both domains. 
To this end, we propose moving all the data points by a controlled amount, so that on the whole, these confusing samples become less ambiguous to classify. The amount and the direction in which the samples are moved must be learnable. For this, we propose to add two fully connected (FC) layers whose outputs we call \emph{transport} terms in the classification (logit) space. These terms represent vectors by which we move a particular sample's representation. That is, within the general formulation of adversarial domain adaptation~\cite{adda}, we propose to replace the external loss by a transport loss using two transport terms $\mathbf{T_1}$ and $\mathbf{T_2}$. We also average the two transported classifier outputs to incorporate sufficient generality. This gives rise to a more specific objective:

\begin{equation}
\begin{aligned}
    \max_D \mathcal{L}_{adv}(G, C, D, \mathbf{T_1}, \mathbf{T_2}) + \mathcal{L}_c(G,C, \mathbf{T_1}, \mathbf{T_2}) \\ + \mathcal{L}_{TL}(G, \mathbf{T_1}, \mathbf{T_2}),
\end{aligned}
\end{equation}

\begin{equation}
\begin{aligned}
    \min_{G,C,\mathbf{T_1}, \mathbf{T_2}} \mathcal{L}_{adv}(G, C, D, \mathbf{T_1}, \mathbf{T_2}) + \mathcal{L}_c(G,C, \mathbf{T_1}, \mathbf{T_2}) \\ + \mathcal{L}_{TL}(G, \mathbf{T_1}, \mathbf{T_2}), 
\end{aligned}
\end{equation}
where $\mathbf{T_1}$ and $\mathbf{T_2}$ are trainable transport vectors, $\mathcal{L}_{TL}$ is the proposed transport loss. We may not include one of the two transport terms in our experiments when there are fewer classes. The objective and the formulation of the transport are explained in more detail next.

\subsection{Transport of samples}

As shown in Figure~\ref{fig:concept1}, when we consider samples  $\mathbf{x}$ and $\mathbf{y}$ from two different classes, it is reasonable to assume that they are somewhat separated in the vanilla classifier space had there been no transport terms. Therefore, maximizing the angles between $\mathbf{T_i}(G(\mathbf{y}))$ (abbreviated as $\mathbf{T_i}(\mathbf{y})$ etc. in the figure) and $\mathbf{T_i}(G(\mathbf{x}))$ for $i\in\{1,2\}$ is expected to drive samples further away from the new decision boundary, which in turn, reduces class confusion. Next, we consider a sample each from the source and the target $\mathbf{x}$ and $\mathbf{x'}$, and let $C(G(\mathbf{x}))$ (abbreviated as $C(\mathbf{x})$ etc. in the figure) and $C(G(\mathbf{x'}))$ be their representations in the classifier (logit, i.e pre-softmax) space. We aim to move them individually by each of the two transports $\mathbf{T_1}$ and $\mathbf{T_2}$ such that the angle between $\mathbf{T_i}(G(\mathbf{x'}))$ and $\mathbf{T_i}(G(\mathbf{x}))$ is maximized for $i\in\{1,2\}$ and the angle between $\mathbf{T_1}(G(\mathbf{x}))$ and $\mathbf{T_2}(G(\mathbf{x}))$ is minimized. Increasing the angles as suggested above can increase the region in which classification boundary between $\mathbf{x}$ and $\mathbf{x'}$ can lie, if indeed these two belong to different classes. On the other hand, if these two samples belong to the same class, their drift apart due to the transport will be compensated for by the much further drift of labeled samples from different classes. This also increases the generality of the classifier in capturing a class.


\subsection{ADA with transport terms}
\label{sec:add}

Adversarial domain adaptation addresses the unsupervised domain adaptation problem by introducing an adversarial objective to the training process. Specifically, the model is trained to simultaneously minimize the classification loss on the source dataset, and to maximize the accuracy on the target dataset. A separate domain classifier is trained to distinguish between the source and target domains. The overall objective should also minimize the thus formulated adversarial transfer loss with respect to the parameters of the discriminator.

Further, the separate domain classifier, i.e. the discriminator, is trained using the following adversarial loss~\cite{cdan,dann}:
\begin{equation} 
    \begin{aligned}
    \mathcal{L}_{adv} =  -\frac{1}{N_s} \sum_{i=1}^{N_s}log(D(G_*(x_i^s)))\\ -\frac{1}{N_t} \sum_{j=1}^{N_t}log(1-D(G_*(x_j^t))),
    \end{aligned}
\end{equation}
where we propose using an interpolation of the two displacements of each sample as follows: 
\begin{equation}
    G_* = \lambda \cdot (C+\mathbf{T_1})(G) + (1 - \lambda) \cdot (C+\mathbf{T_2})(G)
\end{equation}
In Figure~\ref{fig:concept1}, we depict this convex interpolation of $\mathbf{T_1}$ and $\mathbf{T_2}$ as $\mathbf{T}$. We can also rephrase the equation above as:
\begin{equation}
    G_* = \lambda \cdot \mathbf{O_1}(G) + (1 - \lambda) \cdot \mathbf{O_2}(G),
\end{equation}
where, $G$ is the feature extractor, $C$ is the classifier, $\mathbf{T_1}$ and $\mathbf{T_2}$ are the transport layers and $D$ is the discriminator, and $\mathbf{O_i} = C+\mathbf{T_i} $ for $ i \in \{1,2\}$ . The overall objectives are:
\begin{equation} 
    \max\limits_{D} \mathcal{L}_{adv},
\end{equation} 
\begin{equation} 
    \min\limits_{G_*} \mathcal{L}_{c} +\mathcal{L}_{adv},
\end{equation} 
which are done in a two-step optimization process that can be regarded as a mini-max optimization problem.

\subsection{Minimum class confusion}

The minimum class confusion loss $\mathcal{L}_{MCC}$~\cite{mcc} seeks to minimize confusion terms between classes $j$ and $j'$, such that $j \neq j'$ where the indices are exhaustive over the set of classes. On the target domain, the class confusion term between two classes $j$ and $j'$ is given by:  \begin{center}
     $C_{jj'} =  \hat{\mathbf{y}}_{\cdot j}^{\intercal} \hat{\mathbf{y}}_{\cdot j'}^{\intercal}$ 
 \end{center} 
 
A much more nuanced and meaningful formulation of the class confusion would be:

 \begin{equation}
     C_{jj'} = \hat{\mathbf{y}}_{\cdot j}^{\intercal} \mathbf{W} \hat{\mathbf{y}}_{\cdot j'}^{\intercal},
 \end{equation}
where the matrix $\mathbf{W}$ is a diagonal matrix~\cite{mcc}. The diagonal terms $W_{ii}$ given as the softmax outputs of the entropies in classifying a sample $i$. $\hat{\mathbf{y}}_{ij}$ is given as: 
  
 \begin{equation}
  \hat{\mathbf{y}}_{ij} = \frac{ \exp(Z_{ij} / T ) }{\sum_{j'=1} ^ {c} \exp (Z_{ij'}/ T )} ,   
 \end{equation}
where $c$ is the number of classes, $T$ is the temperature coefficient and $Z_{ij}$ is the logistic output of the classifier layer for the class $j$ and the sample $i$. 

After normalizing the class confusion terms, the final MCC Loss function is given as:
 \begin{equation}
     \mathcal{L}_{MCC} = \frac{1}{c} \sum_{j=1}^ {c} \sum_{j' \neq j}^ {c}|C_{jj'}|,
 \end{equation}
which is the sum of all the non-diagonal elements of the class confusion matrix. The diagonal terms represent the "certainty" in the classifier, while the non-diagonal terms represent the "uncertainty" in classification. The MCC loss can be added in conjunction with other domain adaptation methods \cite{mcc}.

\subsection{Transport terms}

Let $B$ denote the batch-size and $c$ denote the number of classes. Let $A = \{a_{ij}\}$ denote the outputs of the first transport vector and $B = \{b_{ij}\}$ denote the outputs of the second transport vector. Then, an estimate of the transport yield $\mathbf{Y}$ using the transport outputs $\mathbf{T_1}$ and $\mathbf{T_2}$ is:
\begin{equation}
\mathbf{Y} = 
\left[\begin{array}{ccc}
a_{1,1} & \cdots  & a_{1,c} \\
a_{2,1} & \cdots & a_{2,c} \\
\vdots  & \ddots & \vdots \\
a_{B,1}  & \cdots & a_{B,c} \\
\end{array}\right]
\left[\begin{array}{ccc}
b_{1,1} & \cdots  & b_{1,c} \\
b_{2,1} & \cdots & b_{2,c} \\
\vdots  & \ddots & \vdots \\
b_{B,1}  & \cdots & b_{B,c} \\
\end{array}\right]^T ,
\end{equation}
which produces a $B \times B$ matrix. This matrix is an example of a bi-linear form. The most general formulation of the loss would be: 
\begin{equation}
    \mathbf{Y} = \mathbf{T_1} \mathbf{C} \mathbf{T_2}^T ,
\end{equation}
where $\mathbf{C}$ is a matrix which includes class information, if the samples belong to the source, and an estimation of the classes (psuedo-labels) if the samples belong to the target. The transport loss $\mathcal{L}_{TL}$ is then defined as: 
\begin{equation}
    \mathcal{L}_{TL} = | \Sigma(\mathbf{Y}) - Tr(\mathbf{Y}) |  ,
\end{equation}
where $\Sigma(\mathbf{Y})$  denotes the sum of all entries of matrix $\mathbf{Y}$ and  $Tr(\mathbf{Y})$ denotes the trace of $\mathbf{Y}$.  $\mathbf{Y}$ can also be written in the following way:\
\begin{equation}
\mathbf{Y} = 
\left[\begin{array}{c}
\mathbf{a_1} \\
\mathbf{a_2}  \\
\vdots  \\
\mathbf{a_B}  \\
\end{array}\right]
\left[\begin{array}{c}
\mathbf{b_1} \\
\mathbf{b_2}  \\
\vdots  \\
\mathbf{b_B}  \\
\end{array}\right]^T .
\end{equation}
The resulting quantity would be a $B \times B$ matrix consisting of dot products. An additional normalizing technique is to scale the matrix entities as cosine similarities. Doing so would equalize the importance of the relationships between all the samples in the batch. In our experiments, this was seen to be favourable for accuracy in some cases but unfavourable in others.  The new loss thus formed is called $\mathcal{L}_{TL_{Cos}}$. 

We also observe that if we use the original loss function, each element of the matrix $\mathbf{Y}$ is weighted by the value of $\mathbf{| a_i | \cdot | b_j |}$ as in the case of $\mathcal{L}_{TL_{Cos}}$, which is reasonable, and takes the length of the transport terms into account. The overall formulation of the total loss function will be: 

\begin{equation}
\label{eq:L_TL}
    \mathcal{L}_{TL} = |\Sigma (\mathbf{T_1} \mathbf{T_2}^T) - Tr(\mathbf{T_1} \mathbf{T_2}^T)| ,
\end{equation}

\begin{equation}
\label{eq:total_loss}
    \mathcal{L}_{total} = \mathcal{L}_{c} + \lambda_{1} \mathcal{L}_{adv} + \lambda_{2} \mathcal{L}_{TL},
\end{equation}
where $\lambda_1$ and $\lambda_2$ are non-negative hyperparameters. Transporting the sample representations in the classification space will be expected to reduce class confusion of the classifier. However, this may cause crowding near some of the decision boundaries in the shifted region. If $\mathbf{M}$ is a matrix that sufficiently captures class confusions in this context, we may reduce the class confusion terms either directly in the bilinear form $\mathbf{Y} = \mathbf{T_1} \mathbf{M} \mathbf{T_2}^T$ so that the transport loss is modified as follows:
\begin{equation}
\label{eq:L_TL_modified}
    \mathcal{L}_{TL} = | \Sigma (\mathbf{T_1} \mathbf{M} \mathbf{T_2}^T) - Tr(\mathbf{T_1} \mathbf{M} \mathbf{T_2}^T) |,
\end{equation}
or separately as a term in the total loss, to accelerate convergence, as follows:

\begin{equation}
    \mathcal{L}_{total} = \mathcal{L}_{c} + \lambda_1 \mathcal{L}_{adv} + \lambda_2\mathcal{L}_{TL} + \mathcal{L}_{MCC}.
\end{equation}
The calculation of the adversarial loss via the discriminator is done as a combination of the soft-max outputs derived from $\mathbf{O_1} = C + \mathbf{T_1}$ and $\mathbf{O_2} = C + \mathbf{T_2}$. The final classification task is done as:
\begin{equation}
    \lambda \cdot \mathbf{O_1} + (1-\lambda)  \cdot \mathbf{O_2}.
\end{equation}
where $\lambda \in (0,1)$. Using a linear combination makes classification less ambiguous after transporting. 

\section{Experiments and Results}

We conducted experiments to benchmark CHATTY on three datasets and ablations studies to understand the contribution of its proposed components. The datasets used for benchmarking CHATTY were Office-Home~\cite{office-home}, Office-31~\cite{office31}, and FHIST~\cite{fhist}. 

The \textbf{Office-Home} dataset is a challenging benchmark dataset, consisting of 15,500 images across 65 classes shared by four extremely distinct domains: Artistic images (Ar), Clip Art (Cl), Product images (Pr), and Real-World images (Rw). All twelve transfer tasks were evaluated using this dataset~\cite{office-home}. 

The \textbf{Office-31} dataset is a widely used benchmark for visual domain adaptation, consisting of 4,110 images across 31 classes from three distinct domains: Amazon (A), Webcam (W), and DSLR (D). All six transfer tasks are evaluated using this dataset~\cite{office31}. 

The \textbf{FHIST} dataset was originally curated for few-shot classification of near-domain target samples~\cite{fhist}, where the source domain is CRC-TP~\cite{crc} data-set and (near-domain) target is NCT-CRC-HE-100K (NCT)~\cite{nct}. It consists of colorectal cancer histology images from two different domains, with 6 classes: Benign, Muscle, Stroma, Inflammatory, Debris and Tumor. For each class, there are close to 20,000 patches in the CRC-TP domain, and around 10,000 patches in the NCT domain. We rephrase this problem in terms of unsupervised domain adaptation, where we do not have access to the target labels. We explore two domain adaptation tasks of adapting the model from CRC-TP to NCT and the inverse problem. For this dataset, we exclude $\mathbf{T_2}$ from the training loop to favour accuracy.

We adopt the standard protocol for unsupervised domain adaptation (UDA) where all labeled source samples and unlabeled target samples are utilized for training. To report our results for each transfer task, we use center-crop images from the target domain and report the classification performance. Our experiments are conducted in PyTorch using the ResNet-50 architecture pre-trained on ImageNet. 

All experiments were done on an NVIDIA RTX 3090 GPU with learning rate = $0.001$ and all graphs were plotted on TensorBoard. Exploratory t-SNE plots were also generated for the source and target samples~\cite{tsne}. The batch size was kept 16 throughout, where 32 samples- 16 from the source and 16 from the target were used to estimate the loss in the training loop.

The weight values $\lambda_2$ used for OfficeHome and Office31 datasets are 0.0002 and 0.0016, respectively, based on the number of classes. 

\subsection{Benchmarking results}

As can be seen in Tables~\ref{table:office-home},~\ref{table:office-31}, and~\ref{table:fhist}, CHATTY with MCC outperforms the previous methods (including SHOT~\cite{shot}, SDAT~\cite{sdat}, f-DAL~\cite{fdal}, and GVB~\cite{gvb})  on all three datasets with comfortable margins, with the exception of SRDC~\cite{srdc} on Office-31.

\begin{table*}
\begin{center}
\begin{tabular}{|l|p{0.7cm}|p{0.7cm}|p{0.7cm}|p{0.7cm}|p{0.7cm}|p{0.7cm}|p{0.7cm}|p{0.7cm}|p{0.7cm}|p{0.7cm}|p{0.7cm}|p{0.7cm}|p{0.5cm}|}
\hline
Method	&	A$\rightarrow$C	&	A$\rightarrow$P	&	A$\rightarrow$R	&	C$\rightarrow$A	&	C$\rightarrow$P	&	C$\rightarrow$R	&	P$\rightarrow$A	&	P$\rightarrow$C	&	P$\rightarrow$R	&	R$\rightarrow$A	&	R$\rightarrow$C	&	R$\rightarrow$P	&	Avg	\\
\hline
ResNet-50~\cite{resnet}	&	34.9	&	50.0	&	58.0	&	37.4	&	41.9	&	46.2	&	38.5	&	31.2	&	60.4	&	53.9	&	41.2	&	59.9	&	46.1	\\
DANN~\cite{dann}	&	45.6	&	59.3	&	70.1	&	47.0	&	58.5	&	60.9	&	46.1	&	43.7	&	68.5	&	63.2	&	51.8	&	76.8	&	57.6	\\
CDAN~\cite{cdan}	&	50.7	&	70.6	&	76.0	&	57.6	&	70.0	&	70.0	&	57.4  &	50.9	&	77.3	&	70.9	&	56.7	&	81.6	&	65.8	\\
MDD~\cite{mdd} &	54.9	&	73.7	&	77.8	&	60.0	&	71.4	&	71.8	&	61.2	&	53.6	&	78.1	&	72.5	&	60.2	&	82.3	&	68.1	\\
GVB-GD~\cite{gvb} &	57.0	&	74.7	&	79.8	&	64.6	&	74.1	&	74.6	&	65.2	&	55.1	&	81.0	&	74.6	&	59.7	&	84.3	&	70.4	\\
SRDC~\cite{srdc} &	52.3	&	76.3	&	81.0	&	69.5	&	76.2	&	78.0	&	68.7	&	53.8	&	81.7	&	76.3	&	57.1	&	85.0	&	71.3	\\
SHOT~\cite{shot} &	56.9	&	\textbf{78.1}	&	81.0	&	67.9	&	\textbf{78.4}	&	\textbf{78.1}	&	67.0	&	54.6	&	81.8	&	73.4	&	58.1	&	84.5	&	71.6	\\
SDAT	~\cite{sdat}&	\textbf{58.2}	&	77.1	&	82.2	&	66.3	&	\underline{77.6} &	76.8	&	63.3	&	\textbf{57.0}	&	82.2	&	74.9	&	\textbf{64.7}	&	\textbf{86.0}	&	72.2	\\
\hline
CHATTY	&	57.7	&	\underline{77.7} &	\underline{82.7} &	\underline{70.7} &	75.4	&	77.1	&	\textbf{71.7}	&	55.4	&	\underline{83.3} &	\underline{77.2} &	59.3	&	85.7	&	\underline{72.8} \\
CHATTY+MCC	&	\underline{57.8} &	77.5	&	\textbf{83.0}	&	\textbf{70.9}	&	76.9	&	\underline{77.8} &	\underline{70.3} &	\underline{55.4} &	\textbf{83.4}	&	\textbf{77.3}	&	\underline{59.4} &	\underline{85.8} &	\textbf{73.0}	\\
\hline
\end{tabular}
\end{center}
\caption{Accuracy (\%) on the Office-Home dataset~\cite{office-home} with 12 different UDA tasks and their average, where all methods were fine-tuned ResNet50~\cite{resnet} pre-trained on ImageNet~\cite{imagenet}.}
\label{table:office-home}
\end{table*}

\begin{table*}
\begin{center}
\begin{tabular}{|l|c|c|c|c|c|c|c|}
\hline
Method	&	A $\rightarrow$ D 	&	A $\rightarrow$ W	&	D $\rightarrow$ W	&	W $\rightarrow$ D	&	D $\rightarrow$ A	&	W $\rightarrow$ A	&	Avg \\
\hline
ResNet-50~\cite{resnet}	&	68.9	&	68.4	&	96.7	&	99.3	&	62.5	&	60.7	&	76.1 \\
DANN~\cite{dann}	&	79.7	&	82.0	&	96.9	&	99.1	&	68.2	&	67.4	&	82.2 \\
CDAN~\cite{cdan}	&	92.9	&	94.1	&	98.6	&	100.0	&	71.0	&	69.3	&	87.7 \\
MDD~\cite{mdd}	&	93.5	&	94.5	&	98.4	&	100.0	&	74.6	&	72.2	&	88.9 \\
GVB-GD~\cite{gvb}	&	95.0	&	94.8	&	98.7	&	100.0	&	73.4	&	73.7	&	89.3 \\
SRDC~\cite{srdc}	&	\textbf{95.8}	&	95.7	&	\textbf{99.2}	&	100.0	&	\textbf{76.7}	&	\textbf{77.1}	&	\textbf{90.8} \\
SHOT~\cite{shot}	&	93.1	&	90.9	&	98.8 &	99.9	&	74.5	&	74.8	&	88.7 \\
f-DAL~\cite{fdal}	&	94.8	&	93.4	&	99.0 &	100.0	&	73.6	&	74.6	&	89.2 \\
\hline
CHATTY	&	91.1	&	94.5	&	\underline{99.3} &	100.0	&	73.7	&	71.7	&	88.4 \\
CHATTY + MCC	&	\underline{95.0}	&	\textbf{95.7}	&	98.6	&	\textbf{100.0} &	\underline{74.7} &	\underline{75.3}&	\underline{89.9} \\
\hline
\end{tabular}
\end{center}
\caption{Accuracy (\%) on the Office-31 dataset~\cite{office31} with 6 different UDA tasks and their average, where all methods are fine-tuned ResNet50~\cite{resnet} pre-trained on ImageNet~\cite{imagenet}.}
\label{table:office-31}
\end{table*}

\begin{table}
\begin{center}
\begin{tabular}{|l|c|c|c|}
\hline
Method	&	CRC $\rightarrow$ NCT	&	NCT $\rightarrow$ CRC	&	Avg	\\
\hline
ResNet-50~\cite{resnet}	&	40.7	&	32.9	&	36.8	\\
DANN~\cite{dann}	&	73.5	&	66.6	&	70.0	\\
CDAN~\cite{cdan}	&	66.2	&	61.4	&	63.8	\\
GVB-GD~\cite{gvb}	&	73.9	&	 \underline{66.7} &	70.3	\\
\hline
CHATTY	&	\underline{77.0} & 64.5	&	\underline{70.7} \\
CHATTY+MCC	&	\textbf{81.6}	&	\textbf{67.9}	&	\textbf{74.7}	\\
\hline
\end{tabular}
\end{center}
\caption{Accuracy (\%) on the FHIST dataset~\cite{fhist} with 2 different UDA tasks and their average, where all methods are fine-tuned ResNet50~\cite{resnet} pre-trained on ImageNet~\cite{imagenet}.}
\label{table:fhist}
\end{table}

\subsection{Effect of MCC and transport losses}

The transport loss is sensitive to small increments after backpropagation. Therefore, the weight of the transport loss must be tuned for every dataset as a hyperparameter. An even higher weight in the total loss might not drive the transport loss to 0, since the updates in the parameters of $\mathbf{T_1}$ and $\mathbf{T_2}$ will overshoot the optimal values. 
In the formulation of the total loss given in Equation~\ref{eq:total_loss} a reasonable conjecture is that the weight of the transport loss should be inversely proportional to the number of classes in the dataset. This was evident from our experiments and also by an analysis of the loss term. We also set $\lambda = 0.5, \lambda_1 = 1$ and $ \mathbf{C} = \mathbf{I}$  (identity matrix) throughout without including the class information intrinsic to the bi-linear form. Instead, we use class information during training by using $\mathcal{L}_{MCC}$ to accelerate the class confusion minimization. 

The best results from previous domain adaptation methods are compared with our approach and the comparison is listed in Table~\ref{table:office-home}.

\begin{figure*}
     \centering
     \begin{subfigure}[b]{0.22\textwidth}
         \centering
         \includegraphics[width=\textwidth]{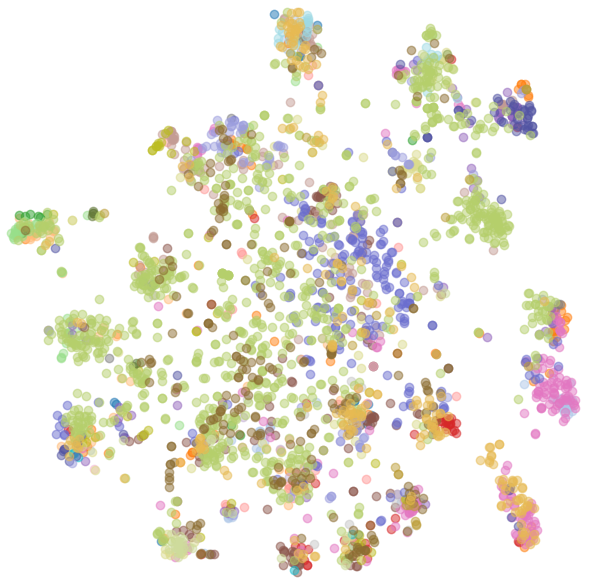}
         \caption{$i = 0$}
         \label{fig:iter0}
     \end{subfigure}
     \hspace{1em}
     \begin{subfigure}[b]{0.22\textwidth}
         \centering
         \includegraphics[width=\textwidth]{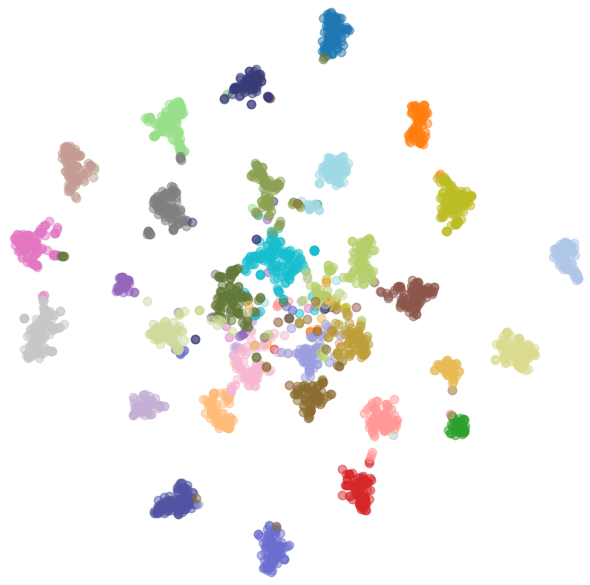}
         \caption{$i = 2500$}
         \label{fig:iter2500}
     \end{subfigure}
     \hspace{1em}
     \begin{subfigure}[b]{0.22\textwidth}
         \centering
         \includegraphics[width=\textwidth]{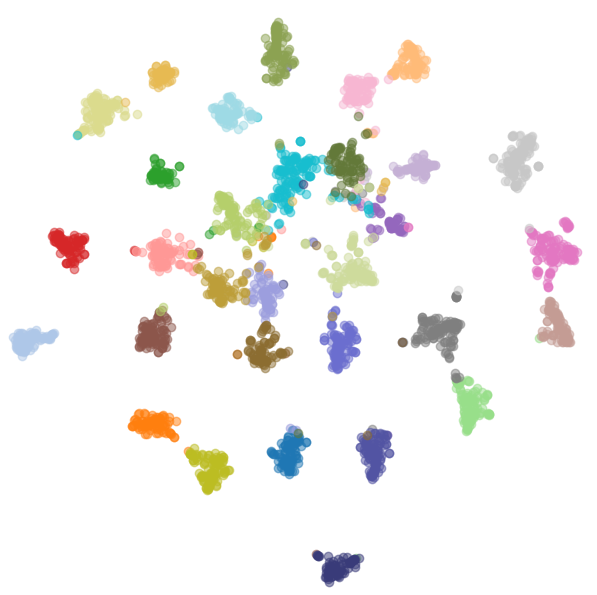}
         \caption{$i = 5000$}
         \label{fig:iter5000}
     \end{subfigure}
     \hspace{1em}
     \begin{subfigure}[b]{0.22\textwidth}
         \centering
         \includegraphics[width=\textwidth]{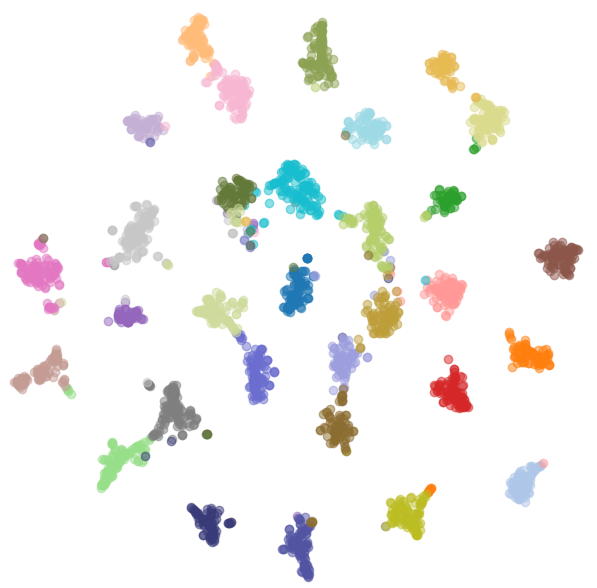}
         \caption{$i = 10000$}
         \label{fig:iter10000}
     \end{subfigure}
        \caption{Two-dimensional representation (using t-SNE \cite{tsne}) of \emph{target} samples after iterations $i = 0,2500,5000$ and $10,000$ of the domain adaptation task D to A on Office-31 shows that the classes initially overlap, but as the training progresses their samples are transported such that they form distinct clusters.}
        \label{fig:tsne}
\end{figure*}

$\mathcal{L}_{MCC}$ directly addresses class confusion of the classifier by minimizing the probabilities of inter-class confusions. However, a careful study of our loss function on Office-Home~\cite{office-home} revealed that our loss addressed class confusion too, by minimizing $\mathcal{L}_{MCC}$, even when $\mathcal{L}_{MCC}$ was not included in the training algorithm.

However, the converse was not true; Minimizing $\mathcal{L}_{MCC}$ did not minimize the $\mathcal{L}_{TL}$, when $\mathcal{L}_{TL}$ was not included in the training loop. We conclude that transporting the samples in the classification space in a nuanced way enables the classifier to make more confident predictions. The very idea of shifting the samples in the classification space seemed to greatly enhance accuracy, and tuning the shifting vectors $\mathbf{T_1}$ and $\mathbf{T_2}$ was but an icing on the cake, which improved the accuracy even further. 

We also suspect that in certain cases, the class confusions are minimized by $\mathcal{L}_{MCC}$ and $\mathcal{L}_{TL}$ synergistically, so that adding the minimum class confusion loss to the total loss caused further improvement in certain domain adaptation tasks. The class confusions that arise due to two samples from different classes being accidentally transported to nearby regions might be the cause of this. 

\begin{figure}
\begin{center}
   \includegraphics[width=1.0\linewidth]{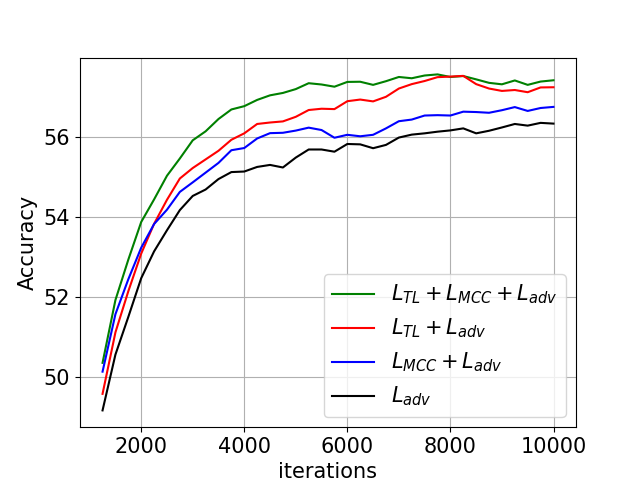}
\end{center}
   \caption{Evolution of accuracy on OfficeHome~\cite{office-home} Ar to Cl domain adaptation task shows that improvements over using the adversarial transfer loss $\mathcal{L}_{adv}$ ~\cite{dann,cdan} can be made by including the minimum class confusion loss $\mathcal{L}_{MCC}$~\cite{mcc}, while further and independent improvements are possible using the proposed transport loss $\mathcal{L}_{TL}$.}
\label{fig:short}
\end{figure}

The transport loss $\mathcal{L}_{TL}$ is sensitive to back-propagating updates, and hence to $\lambda_2$. The sensitivity is reduced by replacing $\mathcal{L}_{TL}$ by $\mathcal{L}_{TL_{Cos}}$ which normalizes every entry in the matrix $\mathbf{Y}$ by the product of the norms. 
Empirically, the best adaptation algorithm was seen to be the loss given by Equation~\ref{eq:L_TL}. When the loss was modified to Equation~\ref{eq:L_TL_modified} it was observed that the convergence was slower than when the class confusion loss was included separately in the total loss. This could be possibly attributed to the back-propagation updates of $\mathbf{M}$ being very slow, owing to the camouflaged nature of $\mathbf{M}$ in $\mathbf{T_1} \mathbf{M} \mathbf{T_2}^T$. Nevertheless, it is interesting to study an optimal choice of $\mathbf{M}$.

In our ablation studies, we observe that CHATTY, when combined with MCC~\cite{mcc} gives the best result on most of the domain adaptation tasks. The evolution of the target classifier accuracy on the domain adaptation from Ar to Cl is plotted in Figure~\ref{fig:short}. We observe that by shifting the samples in the classification (logit) space itself produces a significantly improved accuracy compared to SOTA methods, as shown in Table~\ref{table:office-home}, and tuning the transport vectors increases the accuracy. Furthermore, using the MCC loss in synergy with the transport loss favoured accuracy even more. When using just the transport loss as a means of reducing class confusion, we observe that it does a better job than the MCC loss.

\subsection{Evolution of target domain distribution}

We visualize the movements of 2D nonlinear representation of the \emph{target} domain feature vectors in the classification (logit) space using t-SNE (t-distributed Stochastic Neighbor Embedding)~\cite{tsne} after $0, 2500, 5000,$ and $10000$ training iterations for the D to A task in Office-31. The extent of overlapping between distributions visibly reduces with increase in the number of iterations, as seen in Figure~\ref{fig:tsne}.

\section{Conclusions and Future Directions}

For unsupervised domain adaptation, we propose transporting samples to different regions in the classification (logit) space to allow easier joint classification of samples from source and target domains. Our results show that it is a reasonable approach to minimizing confusions in classifying the target samples in the context of unsupervised domain adaptation, especially when combined with other mechanisms, such as minimizing class confusion.

There is scope of choosing the optimal value of the confusion matrix embedded in the transport loss, by using class information of the source domain samples and an estimate of the label for the target domain samples (pseudo-labels). Doing so would gauge the effective spread of the clusters and lead to more robust classification. Moving two samples of the same class similarly might reduce the class confusion further. Given the multi-modal distribution of the source samples in the classification space, and the multi-modal prior of the target samples, we may approach the problem by introducing an angular shift term, along with linear shifting terms. 
{\small
\bibliographystyle{ieee_fullname}
\bibliography{egbib.bib}
}
 
\end{document}